\documentclass{svjour3}                     
\smartqed  
\usepackage{graphicx}
\usepackage{amsmath}
\usepackage{cite}
\usepackage{url}
\journalname{MISTA 2013}
\begin{document}

\title{Iterated Variable Neighborhood Search for the resource constrained multi-mode multi-project scheduling problem}

\subtitle{Some comments on our contribution to the \emph{MISTA 2013 Challenge}}

\author{Martin Josef Geiger
}

\institute{Martin Josef Geiger \at
              Helmut-Schmidt-University, University of the Federal Armed Forces Hamburg\\
              \email{m.j.geiger@hsu-hh.de}           
}

\maketitle

\section{Introduction}
The resource constrained multi-mode multi-project scheduling problem (RCMMMPSP) is a notoriously difficult combinatorial optimization problem. For a given set of activities, feasible execution mode assignments and execution starting times must be found such that some optimization function, e.\,g.\ the makespan, is optimized. When determining an optimal (or at least feasible) assignment of decision variable values, a set of side constraints, such as resource availabilities, precedence constraints, etc., has to be respected.

In 2013, the \emph{MISTA 2013 Challenge} stipulated research in the RCMMMPSP. It's goal was the solution of a given set of instances under running time restrictions. We have contributed to this challenge with the here presented approach.

\section{Problem description and implications}
In the RCMMMPSP of the \emph{MISTA 2013 Challenge}, several small projects, each of which comprises a set of activities, have to be integrated into a single overall plan. A release date of each project has to be respected, and resources that are jointly used by several projects may exist. Two types of resources exists: renewable and non-renewable. The latter are always specific to each project, i.\,e.\ they are never shared among the projects.

The quality of the obtained schedule is evaluated by two objective functions, namely the minimization of the total project delay (TPD), and the total makespan (TMS), which are considered in a lexicographical ordering with TPD over TMS.

For a more detailed description of the RCMMMPSP, we refer to \url{http://allserv.kahosl.be/mista2013challenge/}.

\subsection{A remark on the multi-project characteristic}
It is possible to combine the set of projects into an overall super-project by relabeling the numbering of the activities. Quite intuitively, this polynomial-time transformation allows for a rather straight-forward representation of alternatives, in which each activity is represented by a unique index $i, i  = 1, \ldots, n$ (see section~\ref{sec:representation} below). However, we cannot rule out the possibility that a separate treatment/ optimization of the different small sub-projects is not of any potential value.

\subsection{\label{sec:mode:assignment}A note on the mode assignment}
In the general case, i.\,e.\ the case of at least two non-renewable resources, the determination of a feasible mode assignment is $\mathcal{NP}$-complete \cite{kolisch:1997:article}. This means that, depending on the datasets, this sub-aspect of the problem may pose a problem on it's own. Any heuristic approach for selecting mode assignments is therefore risky, especially under running time restrictions.

\subsection{Two remarks on the objective functions}
It can be shown that both objectives are \emph{regular} functions. It is therefore possible to restrict the computations to active plans only, without excluding the optimum. This is an important observation, that allows the use of a serial scheduling scheme (SSS) \cite{kelley:1963:incollection,demeulemeester:2002:book}. In such a scheme, a schedule is constructed by sequentially assigning activities into a schedule, one activity at a time. An earliest-possible-(`left-shifted-')scheduling-rule may here be employed for obtaining an active schedule. This implies that an optimization approach can then search the sequence in which the activities are placed in the final schedule, instead of searching the activities' starting times themselves.

Note that, in the general case, a combination of the two criteria into a single evaluation function is not possible. If however we may safely assume that, for a particular set of problem instances (datasets), TMS $< \alpha$, then a combination of TPD and TMS is possible such that: $F$ = $\alpha$ TPD $+$ TMS. In the context of the \emph{MISTA 2013 Challenge}, the organizers guarantee a value of $\alpha = 100{,}000$.

\section{Solution approach}
\subsection{\label{sec:representation}Representation of alternatives}
An alternative $\vec{X}$ is associated with two vectors $\vec{M}=(m_1, \ldots, m_n)$ and $\vec{S}=(s_1, \ldots, s_n)$. $\vec{M}$ encodes for each activity $i$ the choice of the execution mode $m_i$, and $\vec{S}$ is a permutation of the activity indexes. In the following, let $\mathcal{X}$ denote the set of feasible alternatives.

While the information in $\vec{M}$ is obvious, $\vec{S}$ can be decoded into a schedule by a SSS. In this transformation, the feasibility of $\vec{X}$ with respect to $\vec{S}$ is maintained at all times as we adhere to the precedence constraints of activities while carrying out the procedure. Scheduling conflicts are solved such that priority is given in accordance with the sequence of activities in $\vec{S}$. Using such a SSS implies that any permutation of activity indexes leads to a feasible schedule.

\subsection{Generation of first feasible alternatives}
The generation of initial alternatives is twofold:
\begin{enumerate}
\item First, a random mode assignment $\vec{M}$ is computed. If $\vec{M}$ is infeasible, i.\,e.\ by exceeding the capacity of the non-renewable resources, a repair procedure is triggered, which randomly alters values $m_i$ in $\vec{M}$ until the feasibility of the mode choices is restored. After a maximum number of unsuccessful repair attempts, $\vec{M}$  is randomly reconstructed from scratch. Note that the repair procedure is guided towards a feasible mode assignment by minimizing the overall excessive use of the non-renewable resources (this assumes a value of $0$ for a feasible $\vec{M}$).

In the light of section~\ref{sec:mode:assignment}, such a procedure could turn out to be problematic, as it does not guarantee the identification of the feasible mode assignment. It has therefore been tested on all 1,642 resource constrained multi-mode project scheduling datasets which were delivered in the \emph{MISTA 2013 Challenge}. In conclusion, the procedure repeatedly and consistently managed to create feasible mode vectors for all instances.

\item Second, a scheduling sequence $\vec{S}$ is constructed. The SSS described above is used here. For the first alternative, scheduling conflicts are solved by giving priority to activities with smaller lower bounds on their possible starting times.
\end{enumerate}

\subsection{Neighborhood search}
The initially constructed alternative $\vec{X}$ is then improved by means of local search. The proposed concept is based on the ideas of Variable Neighborhood Search \cite{hansen:2001:article} (VNS) and Iterated Local Search \cite{lourenco:2003:incollection}. Starting from an initial (feasible) alternative, neighboring alternatives are generated by means of several neighborhood operators and tested for acceptance. Once an alternative is reached that cannot be improved by any operator (a local optimum), a (possibly worsening) perturbation move is performed and search continues from this alternative.

Four different neighborhood operators have been implemented:
\begin{enumerate}
\item Exchange (EX), which exchanges the position of two activities in $\vec{S}$. We denote the set of the neighbors of $\vec{X}$ by the operator EX as EX($\vec{X}$). A neighboring solution $\vec{X'} \in$ EX($\vec{X}$) is accepted iff $F(\vec{X'}) < F(\vec{X})$.

\item Inversion (INV), which inverts a subsequence of activities within $\vec{S}$. We accept $\vec{X'} \in$ INV($\vec{X}$) iff $F(\vec{X'}) < F(\vec{X})$.

\item Single mode change (SMC), which changes the mode assignment of a single element in $\vec{M}$. $\vec{X'} \in$ SMC($\vec{X}$) is accepted iff $F(\vec{X'}) < F(\vec{X}) \wedge \vec{X'} \in \mathcal{X}$.

\item Two modes changes (TMC), which changes the mode assignments of two elements in $\vec{M}$. We accept $\vec{X'} \in$ TMC($\vec{X}$) iff $F(\vec{X'}) \leq F(\vec{X}) \wedge \vec{X'} \in \mathcal{X}$. Note that this acceptance rule differs from the ones of EX, INV, and SMC: Accepting alternatives of equal quality appears to have a beneficial effect for this operator, possibly due to some diversification effect that results from it.
\end{enumerate}

In case of smaller instances ($n<307$: this value has been determined by a series of experiments), each neighborhood is repeatedly searched until no improvement can be identified. Then, the next neighborhood operator is taken. For larger instances, this does not appear to be an appropriate strategy. Here, we search each neighborhood, and then move directly to the next one, thus iterating through the set of neighborhood operators more quickly.

Escaping local optima: The perturbation move consists in randomly changing the execution mode of a single activity. This move is followed by calling the randomized mode-repair-procedure, which, if required, alters several other mode assignments and thus implicitly provides some sort of diversification.

\subsection{A note on the parallelization of the approach}
In case of smaller instances ($n<307$), the above presented local search is simply executed in parallel (one master thread, four worker threads). Once a worker thread reaches a local optimum, the currently best found alternative is updated in the master thread and search continues with this best known alternative (perturbation + VNS).

Larger instances come with the difficulty of obtaining a local optimum in the first place. Therefore, all available CPU-cores are concentrated on a single solution by dividing up the investigation of a neighborhood among them.

\section{Experiments and results}
\subsection{Implementation details and hardware}
Our algorithm is coded using Microsoft Visual Studio 2012. The program therefore requires a Microsoft ``Windows'' operating system and a current version of the ``.NET Framework'' (version 4.5). We have used an Intel X5550 processor, running at 2.67 GHz for the test runs. Despite the availability of eight cores on our machine, only four threads of the local search algorithm are executed in parallel. The target platform for the implemented code optimizations is 64-bit (OS/CPU).

The benchmark program of the 2011 International Timetabling Competition gave, on the organizers machine, which is an Intel Core i7-2600 (3.4 GHz), an execution time of 619 seconds. On our Intel X5550, the very same program reported a running time of 804 seconds (average of 5 trials). We have concluded that the Intel X5550 should be given 29.9\% more computing time, resulting in a permitted computing time of 389 seconds for each run. Note that the comparison of the ITC 2011 benchmark program pretty much reflects the different clock speeds of the CPUs.

\subsection{Results and execution of the program}
Table~\ref{tbl:results} reports the results of 20 test runs, i.\,e.\ the quality of the best found alternatives and the median values. We can observe a considerable deviation of the best value from the median. This indicates that the approach is not consistently solving the datasets each and every time it is executed. While this is typical for a stochastic solver, especially under, for the larger datasets, tight running time restrictions, we suspect that there is still some room for improving the approach.

\begin{table}
\caption{\label{tbl:results}Best results and median values over 20 test runs}
  \begin{tabular}{lrr|rr|lrr|rr}
  \hline
  & \multicolumn{2}{c}{Best results} & \multicolumn{2}{c|}{Median}&&\multicolumn{2}{c}{Best results} & \multicolumn{2}{c}{Median}\\
    Dataset & TPD & TMS & TPD & TMS &  Dataset & TPD & TMS & TPD & TMS\\
    \hline
{\tt A-1.txt}&	1&	23&	1&	23&	{\tt B-1.txt}&	349&	130&	357&	131\\
{\tt A-2.txt}&	2&	41&	2&	41	&{\tt B-2.txt}&	481&	171&	515&	178\\
{\tt A-3.txt}&	0&	50&	0&	50	&{\tt B-3.txt}	&604&	214&	624&	216\\
{\tt A-4.txt}&	65&	42&	65&	42	&{\tt B-4.txt}	&1283&	287&	1328&	286\\
{\tt A-5.txt}&	153&	104	&155	&106&	{\tt B-5.txt}&	866&	252&	909&	265\\
{\tt A-6.txt}&	144&	94	&150	&97	&{\tt B-6.txt}&	1067&	246&	1098&	250\\
{\tt A-7.txt}&	601&	206	&620	&204&	{\tt B-7.txt}&	827&	232&	844&	241\\
{\tt A-8.txt}&	319&	162	&340	&161&	{\tt B-8.txt}&	3618&	565&	3729&	574\\
{\tt A-9.txt}&	225&	128	&240	&131&	{\tt B-9.txt}&	4606&	783&	4825&	806\\
{\tt A-10.txt}&	920&	313	&955	&321&	{\tt B-10.txt}&	3541&	473&	3751&	474\\
\hline
  \end{tabular}
\end{table}

Reproduction of the reported results should be possible by re-running the program. The command line works as follows:

\centerline{{\tt npuScheduler.exe A-1.txt A-1\_sol\_13264537.txt 300 13264537}}

The program {\tt npuScheduler.exe} is run on instance {\tt A-1.txt} (which has to be stored in the very same directory as the executable). A solution file, {\tt A-1\_sol\_13264537.txt}, is produced after 300 seconds (in the same directory as the application {\tt npuScheduler.exe}), using a random number seed of {\tt 13264537}.

\vspace{-0.3cm}
\section{Limitations and future research}
The proposition of the solution approach assumes that the datasets provided by the organizers of the \emph{MISTA 2013 Challenge} are a representative subset for the problem that should be solved. This implies the following limitations of our concept:
\begin{itemize}
\item The generation of feasible mode assignments by means of a randomized algorithm can, as described above, be a problematic issue in case of very limited non-renewable resources. In such situations, infeasible mode selections could be (temporarily) accepted in the search procedure, and evaluated by means of a penalty function \cite{hartmann:2001:article}.

\item The implemented neighborhood operators are, with the exception of the SMC, of complexity $\mathcal{O}(n^2)$. In can be expected that larger instances will call for adaptations of the neighborhood search.

In the \emph{MISTA 2013 Challenge}, the largest test instance was {\tt B-9} with $n = 642$ (incl.\ dummy-activities). For this particular dataset, no local optimum was found within the given time limit, and thus, no perturbation move was performed here.
\end{itemize}
Naturally, future research should address the above mentioned limitations.

\vspace{-0.3cm}
\bibliographystyle{spmpsci}      
\bibliography{literature}   

\end{document}